# CE-RS-SBCIT: A Novel Channel-Enhanced Hybrid CNN–Transformer with Residual, Spatial, and Boundary-Aware Learning for Brain Tumor MRI Analysis

Mirza Mumtaz Zahoor[1], Saddam Hussain Khan[2]*

[1]Faculty of Computer Sciences, Ibadat International University, Islamabad, 44000, Pakistan

[2]Artificial Intelligence Lab, Department of Computer Systems Engineering, University of Engineering and Applied Sciences (UEAS), Swat 19060, Pakistan

**Email:** saddamhkhan@ueas.edu.pk

## Abstract

Brain tumors remain among the most lethal human diseases, where early detection and accurate classification are critical for effective diagnosis and treatment planning. Although deep learning-based computer-aided diagnostic (CADx) systems have shown remarkable progress. However, conventional convolutional neural networks (CNNs) and Transformers face persistent challenges, including high computational cost, sensitivity to minor contrast variations, structural heterogeneity, and texture inconsistencies in MRI data. Therefore, a novel hybrid framework, CE-RS-SBCIT, is introduced, integrating residual and spatial learning-based CNNs with transformer-driven modules. The proposed framework exploits local fine-grained and global contextual cues through four core innovations: (i) a smoothing and boundary-based CNN-integrated Transformer (SBCIT), (ii) tailored residual and spatial learning CNNs, (iii) a channel enhancement (CE) strategy, and (iv) a novel spatial attention mechanism. The developed SBCIT employs stem convolution and contextual interaction transformer blocks with systematic smoothing and boundary operations, enabling efficient global feature modeling. Moreover, Residual and spatial CNNs, enhanced by auxiliary transfer-learned feature maps, enrich the representation space, while the CE module amplifies discriminative channels and mitigates redundancy. Furthermore, the spatial attention mechanism selectively emphasizes subtle contrast and textural variations across tumor classes. Extensive evaluation on challenging MRI datasets from Kaggle and Figshare, encompassing glioma, meningioma, pituitary tumors, and healthy controls, demonstrates superior performance, achieving 98.30% accuracy, 98.08% sensitivity, 98.25% F1-score, and 98.43% precision. These results underscore the potential of CE-RS-SBCIT as a robust diagnostic framework that bridges CNN and transformer paradigms for reliable and efficient brain tumor analysis.

# 1. Introduction

The brain, the body's most complex and vital organ that regulates the nervous system, can develop brain tumors (BTs) from abnormal and uncontrolled cell growth. BTs are among the deadliest malignancies and the most common type of cancer worldwide, according to statistics from Global Cancer Statistics [1]. In modern clinical imaging, early detection and classification of BTs are critical research priorities because they directly affect patient survival and the effectiveness of treatment planning [2]. Accurate diagnosis is therefore essential and relies on multimodal imaging and more advanced machine-learning (ML) systems, which enable comprehensive characterization of tumor morphology and phenotype [3].

Researchers and scientists have developed advanced techniques for identifying BTs. Commonly used methods to detect irregularities in the shape, size, or position of brain tissue include MRI and computed tomography (CT). Among these, MRI is the preferred imaging technique among physicians and researchers because it provides detailed information about the shape, location, and size of BTs in medical images (MIs). Furthermore, it is generally considered a faster, more cost-effective, and safer option. Manually assessing brain MRI scans is challenging for radiologists when it comes to accurately identifying and classifying BTs from MIs. To ease this burden and assist radiologists and clinicians in evaluating MIs, the computer-aided diagnosis (CADx) system is essential.

Various research areas in MI analysis are currently being explored, comprising domains such as identification, recognition, and segmentation in medical imaging [4], [5], [6], [7], [8], [9], [10]. Traditional ML methods depend on manually formed features that demand substantial prior knowledge, such as identifying the tumor's location in a medical scan. However, this approach poses a significant risk of human error, limited reproducibility, and poor adaptability across heterogeneous imaging modalities. Furthermore, manual feature engineering often fails to capture complex nonlinear patterns, is highly sensitive to noise, and struggles with limited or imbalanced MRI datasets containing brain abnormalities and other organ irregularities. Consequently, developing automated frameworks capable of integrating both low- and high-level representations without human intervention has become critical. As BTs datasets expand, the demand for robust and scalable feature extraction strategies continues to grow [11], [12], [13], [14], [15].

Convolutional Neural Networks (CNNs) have emerged as powerful tools in this direction, excelling at learning spatial hierarchies and localized feature representations, which underpin their dominance in medical image classification and segmentation [16]. Their convolutional kernels effectively capture fine-grained textures and local structures, making them particularly effective in tumor detection and region-based analysis. However, CNNs inherently struggle to model long-range dependencies and global contextual relationships, which are vital for holistic medical image interpretation. Vision Transformers (ViTs), leveraging self-attention, address this limitation by modeling inter-regional dependencies and emphasizing salient regions across the entire image. Nonetheless, their reliance on large-scale datasets and intensive computational resources restricts applicability in data-constrained medical imaging scenarios [17]. Hybrid CNN–ViT architectures, therefore, offer a promising solution by combining CNNs' strength in local feature extraction with ViTs' capacity for global context modeling, enabling more comprehensive and reliable medical image representations [18].

Earlier hybrid approaches predominantly extended CNNs with additional convolutional layers or incorporated attention mechanisms to enhance spatial learning [19]. Dosovitskiy et al. [20] introduced the ViT, which represents images as patch sequences and processes them via self-attention. While ViTs demonstrated competitive performance, they suffer from quadratic computational complexity and require large-scale datasets for stable training. Liu et al. [21] addressed complexity with the Swin Transformer, introducing hierarchical self-attention within non-overlapping windows to reduce cost to linear scaling. However, large annotated datasets remain a bottleneck. Touvron et al. [22] developed the Data-efficient Image Transformer (DEiT), employing knowledge distillation to reduce data dependency, though quadratic complexity persists. Similarly, Tolstikhin et al. [23] designed the MLP-Mixer, which eliminates both convolutions and self-attention by relying on token and channel mixing; however, accuracy degraded compared with ViTs. Pengyuan et al. [24] reported SCViT, incorporating lightweight channel attention to capture long-range dependencies, yet dataset size and computational complexity remained challenges. In medical imaging, Bazi et al. [25] leveraged ViTs with aggressive augmentation strategies like cut-mix, mix-up to partially address data scarcity, but high computational overhead persisted. While such methods improved local feature extraction, they remained inadequate in capturing global dependencies. Conversely, attention-based hybrids emphasized salient regions but lacked CNNs' robustness in modeling localized spatial hierarchies. Subsequently, prior designs often favored one paradigm over the other, limiting their complementarity. To bridge these gaps, this study proposes a customized CNN–ViT framework with an integrated boosting mechanism for brain MRI analysis, effectively combining CNNs' local feature learning with ViTs' capacity for global dependency modeling [26]. The proposed framework enhances representational richness, improving robustness in data-limited settings, and optimizing BTs analysis performance. The proposed work offers the following contributions:

- A Novel hybrid BT diagnostic framework named "CE-RS-SBCIT is developed, integrating CNNs and ViT to jointly capture global contextual dependencies and local fine-grained details in brain tumor BT MRI images. The proposed framework integrates four heterogeneous yet complementary modules: SBCIT, residual learning CNN, spatial learning CNN, and channel enhancement (CE).

- The Smoothing and Boundary-based CNN-Integrated Transformer (SBCIT) introduces a generic STEM CNN with customized CNN-integrated Transformer (CIT) blocks, enhanced by systematic smoothing and boundary (SB) operations. The customized CIT employs multi-head attention to develop pixel-level contextual interactions, while a lightweight attention block improves computational efficiency and facilitates effective local feature retrieval.

- Enriched feature space through integration: In CE-RS-SBCIT, SBCIT channels are enhanced with auxiliary features generated by transfer learning (TL)-based residual and spatial CNN blocks. These residual CNNs employ inverse residual concepts to ensure efficient gradient propagation in deep structures, while spatial CNNs extract discriminative textural and contrast features. This integration forms a diverse and

- enriched feature space that strengthens representational capacity for tumor heterogeneity.
- A novel spatial attention mechanism is incorporated to refine pixel-level selection, emphasize subtle inter-class variations, and suppress redundancy. Comprehensive experiments on publicly available Kaggle and Figshare datasets, including glioma, meningioma, pituitary tumors, and healthy controls, demonstrate that CE-RS-SBCIT achieves superior diagnostic performance compared with recent CNN and ViT models.

The study is structured as follows: Section 2 provides a survey of research precedents, Section 3 proposes a new classification framework specific to BTs. Section 4 outlines the dataset, implementation, and evaluation metrics. Section 5 reports on results exploration, an ablation study, and comparative analysis overall, and Section 6 provides a conclusion and outlines future research directions.

## 2. Related Studies

Deep learning (DL) has become the dominant paradigm in MIs analysis, particularly for BTs classification using MRI scans. DL integrates both feature learning and classification within a single end-to-end pipeline, requiring minimal preprocessing [27], [28], [29]. CNNs have been the most widely employed architecture, demonstrating the strong ability to bridge the semantic gap between low-level imaging features and high-level expert interpretations [30]. Numerous CNN-based studies have reported promising results across diverse BT datasets.

Previously, a CNN model in [31] achieved 91.4% accuracy using a publicly available dataset, while Sajjad et al. [32] improved performance to 94.58% through data augmentation and fine-tuning with Softmax classification. Similarly, Ahmet and Muhammad [33] fine-tuned ResNet-50 by replacing its final layers, attaining 97.2% accuracy. Sultan et al. [34] validated their CNN framework on two datasets: one comprising 3,064 MRI scans of glioma, meningioma, and pituitary tumors (96.13% accuracy) and another with 516 scans across tumor grades (98.7% accuracy). Özyurt et al. [35] combined AlexNet features with an SVM classifier after NS-EMFSE-based segmentation, yielding a 95.62% detection rate. Using the BRATS dataset, Sajja et al. [36] classified malignant and benign tumors with VGG16, reaching 96.7% accuracy. Despite these successes, CNNs remain limited by local receptive fields, inductive bias, and restricted ability to model global dependencies, often leading to sensitivity to textural heterogeneity and contrast variations. To overcome these shortcomings, Transformer-based models have recently been explored in BT classification.

A deep ViT has been trained from scratch for multi-class classification of gliomas, meningiomas, and pituitary tumors using T1-weighted MRI sequences, with interpretability supported by Grad-CAM, LIME, and SHAP, achieving 91.61% accuracy compared to 83.37% with CNNs, though requiring substantial data and computational resources for scalability [37]. To address these limitations, a Transformer-Enhanced CNN combined CNN modules for local features with Transformer blocks for global dependencies, attaining 96.75% accuracy on the BraTS 2018 dataset, but raising concerns regarding generalizability across heterogeneous cohorts [38]. Further, a relative position encoding strategy and residual structures within MLP layers were introduced to enhance spatial learning and training stability, yielding 91.36% accuracy on an augmented open-source dataset, yet highlighting trade-offs between architectural complexity and computational efficiency [39].

Finally, CNN-based models offer strong performance on moderate-sized datasets but are inherently constrained in modeling the global context. Conversely, transformer-based methods excel in capturing long-range dependencies but are hampered by high data and computational requirements. Moreover, most studies rely on TL from natural image pre-training, which neglects domain-specific patterns in BT MRI data, thereby limiting diagnostic precision. These gaps highlight the demand for hybrid models that integrate CNN-based local feature extraction with ViT-driven global dependency modeling, while improving efficiency and scalability for clinical use. Additional limitations are outlined below.

- CNNs generally focus on local information, which can reduce spatial correlations and restrict their ability to capture more global and complex patterns.
- ViTs divide images into linear patches, making them sensitive to patch size and potentially inadequate in extracting fine-grained local details or low-resolution features.

- Additionally, DL model training is often impeded by the vanishing gradient problem, where gradients diminish across layers, limiting learning in deeper architectures.

## 3. Proposed Methodology

This study introduces a novel BT classification framework for MRI that integrates CNN with ViT-based techniques. To enhance generalization and reduce potential dataset bias, augmentation is applied to the training samples during preprocessing. The proposed framework is examined through three experimental configurations. The proposed BT classification framework involves three experimental setups: (1) CE-RS-SBCIT, (2) an assessment of standard ViTs and hybrid CNN-ViT, and (3) an evaluation of existing CNNs. The discriminative capability of the proposed classification technique is empirically evaluated by various typical performance measures, and the results are assessed by evaluating them with existing ViTs/CNNs, as shown in Figure 1.

### 3.1 Preprocessing

The preprocessing phase was designed to enhance model learning, improve generalization, and reduce overfitting. The dataset suffers from class imbalance, which can bias predictions and increase false positives. To address this, a systematic data augmentation (DA) strategy was applied to expand training diversity and improve minority class representation. DA included controlled geometric transformations such as rotation, scaling, shearing, translation, and reflection, applied within parameter ranges detailed in Table 1. These operations were carefully constrained to preserve anatomical consistency while introducing realistic variability in tumor morphology and positioning. The DA pipeline was implemented using the Keras image augmentation library, enabling real-time generation of diverse training samples. This ensured consistent exposure to a wide range of tumor appearances, mitigating imbalance effects and enhancing the robustness of the proposed MRI-based classification framework.

**Table 1.** Data Augmentation detail.

| Perameters | Values |
|---|---|
| Rotation | [± 3] |
| Shearing | [0 , 30] |
| Scalling | [1, 1.5] |
| Translation | ±5 |
| X-Y Reflection | ±1 |

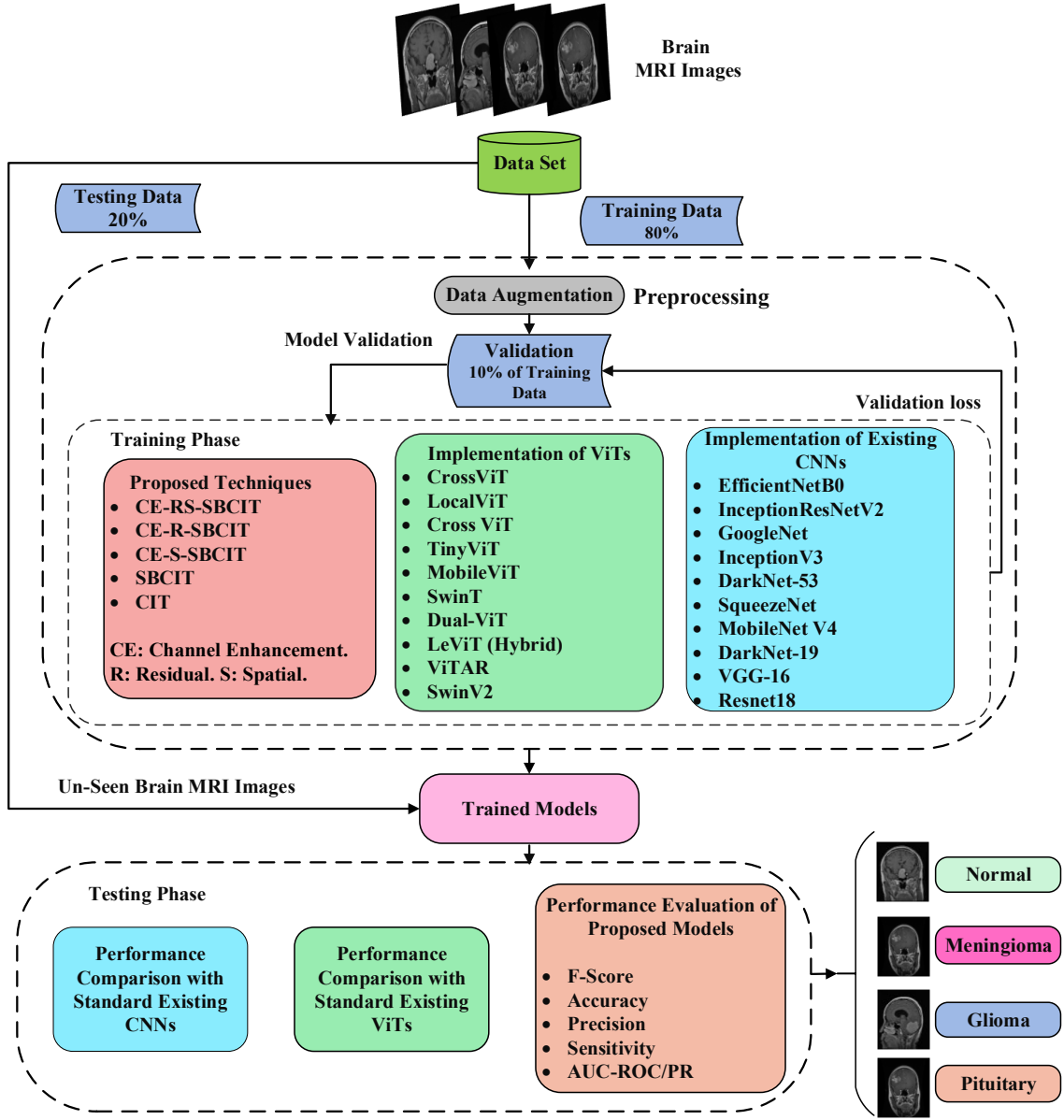

**Figure 1.** Block design of the proposed CE-RS-SBCIT brain tumor MRI images classification framework.

## 3.2 The Developed CE-RS-SBCIT Technique

The proposed CE-RS-SBCIT model integrates a novel SBCIT backbone with TL-based residual and spatial learning CNN blocks, designed for auxiliary feature map generation and concatenation. Recent advances in hybrid ViT-CNN architectures have demonstrated strong potential in MI analysis, motivating the development of SBCIT. In ViTs, images are divided into non-overlapping linear patches that are propagated through encoder blocks via linear projections. However, these projections often limit the model's ability to preserve fine-grained local and structural information due to their inherent linearity. Conversely, CNNs excel at capturing local, translation-invariant, and spatially correlated patterns within two-dimensional neighborhoods, whereas ViTs are more effective in modeling long-range dependencies and global contextual relationships. To reconcile these complementary strengths, the SBCIT module is introduced to enhance convolutional effectiveness at the early processing stages of BT MRI. This is achieved through an optimized patching and tokenization strategy that enriches local feature extraction while maintaining global context representation.

The CE-RS-SBCIT backbone comprises four hierarchical stages, each combining the SBCIT module with residual and spatial learning blocks. Convolutional operations in this pipeline strengthen the model's ability to extract discriminative features, while the residual and spatial learning modules contribute to robust, multi-level feature representations. These diverse features, derived from both the SBCIT module and the CNN-based residual and spatial blocks, are concatenated to construct a comprehensive and complementary feature space at the target level. Furthermore, the CE-RS-SBCIT framework incorporates an attention mechanism that emphasizes subtle discriminatory cues and enhances intra-class variance modeling, thereby improving diagnostic precision for BT MRI classification. Figure 2 illustrates the detailed architecture of SBCIT and demonstrates the integration of residual and spatial learning modules within the overall CE-RS-SBCIT network.

### 3.2.1 The Proposed SBCIT Architecture

Our study intends to develop a hybrid system that integrates the capabilities of CNNs and ViTs. The developed SBCIT starts with a STEM CNN for initial image processing, followed by customized CIT blocks. Additional CNN layers are used to extract both smoothing and boundary features. The architecture processes images using the STEM CNN, then divides them into patches, and embeds these patches as tokens. These tokens are then leveraged for global feature extraction inside the transformer blocks (Figure 2). The CNN block begins with a 3x3 convolutional layer (Conv-L) with a stride of 2 and 64 output channels. This is pursued by two additional 3x3 convolutions with a stride of one, which optimize the prominent local feature extraction.

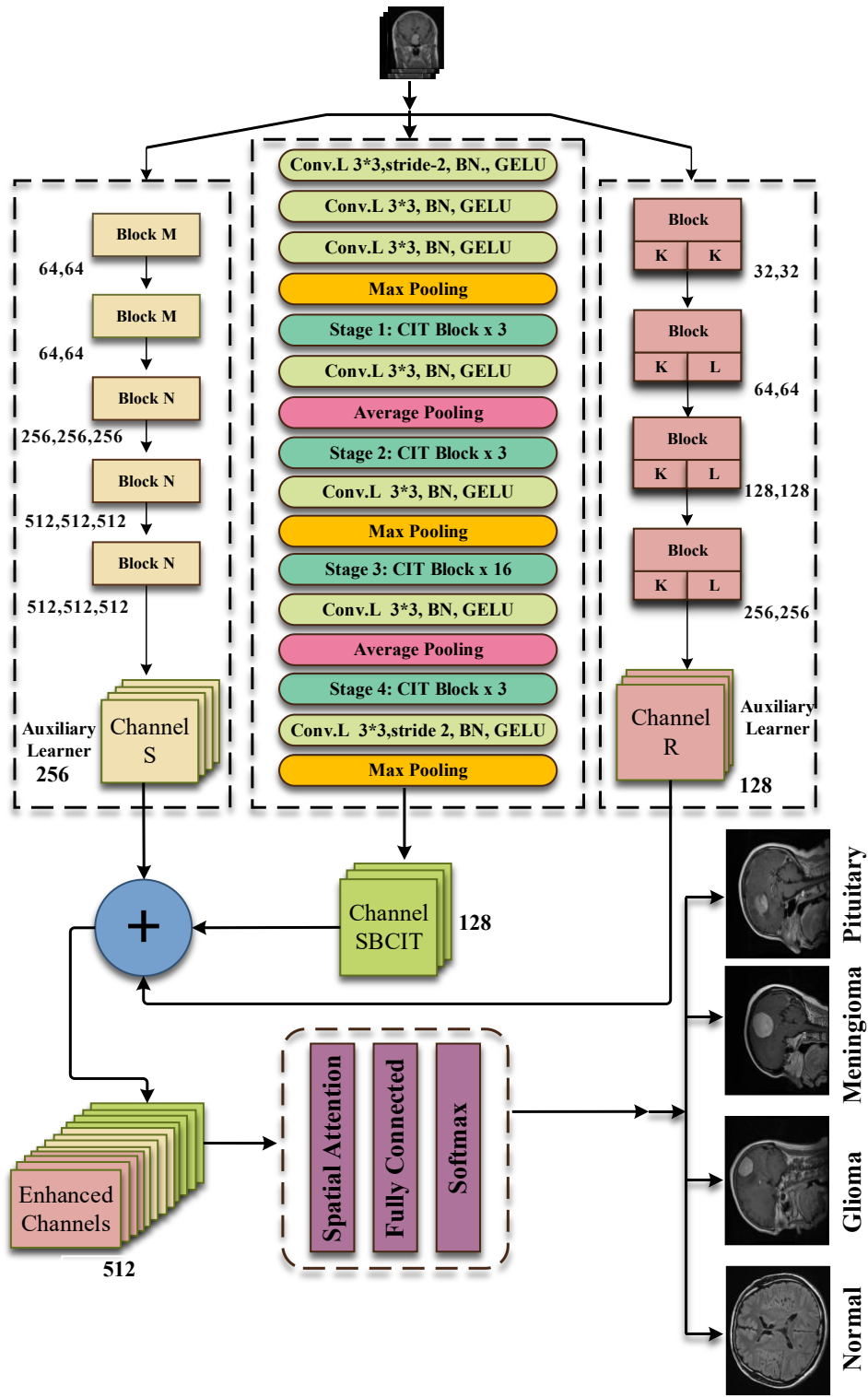

**Figure 2.** Block diagram of the developed CE-RS-SBCIT model.

The SBCIT architecture is structurally customized and consists of four stages, each aimed at generating feature maps at various scales for accurate prediction in complex tasks. After every stage, convolutional layers (Conv-L) with activation functions are employed, alongside average and max pooling methods to improve regional homogeneity and combine structural learning, as represented in Eqs. (1-3). The first convolutional block expands the input feature map to 64 channels, with channel depth progressively increasing after each pooling stage. This systematic amplifying of feature dimensions facilitates the effective downsampling of intermediate feature resolutions after each stage, leading to feature invariance, a more robust model, and better generalization. After the convolutional stages, a patch embedding layer transforms the feature maps into tokens of various dimensions. Ultimately, the SBCIT architecture generates four hierarchical feature channels at different resolutions, resembling the multi-scale output format found in conventional CNNs.

$$\mathbf{x}_{q,r} = \sum_{i=1}^{m} \sum_{j=1}^{n} x_{q+i-1, r+n-1} \, f_{i,j} \tag{1}$$

$$\mathbf{x}^{max}_{q,r} = max_{i=1,\dots,w, j=1,\dots,w} x_{q+i-1, r+j-1} \tag{2}$$

$$\mathbf{x}^{max}_{q,r} = \frac{1}{w^2} \sum_{i=1}^{m} \sum_{j=1}^{n} x_{q+i-1, r+j-1} \tag{3}$$

The Conv-L feature map, denoted as '$x$' with size '$q \times r$', is defined in Eq. (1) using a filter '$f$' of dimension '$i \times j$', producing an output range of [1 to q–m+1, r–n+1]. The convolved output $x_q$ is processed using a homogeneous structural frame size as '$w$' in Eqs. 2-3.

### 3.2.2 Developed CIT Block

The proposed CIT architecture consists of four systemically designed stages, each intended to generate feature channels at different scales to address the prediction challenges. Within each stage, CIT blocks are methodically stacked to facilitate feature transformation while maintaining the input resolution. The integration of CIT blocks within each stage is illustrated in Figure 3. A 3×3 Conv-L with a normalization layer (NL) forms the patch embedding at each phase, producing hierarchical representations in the CIT framework. In addition, the customized CIT architecture enables the creation of multi-scale representations, which are crucial for complex tasks. Various CIT configurations, along with the use of stridden Conv-L applied to the input, offer distinct advantages for BT analysis.

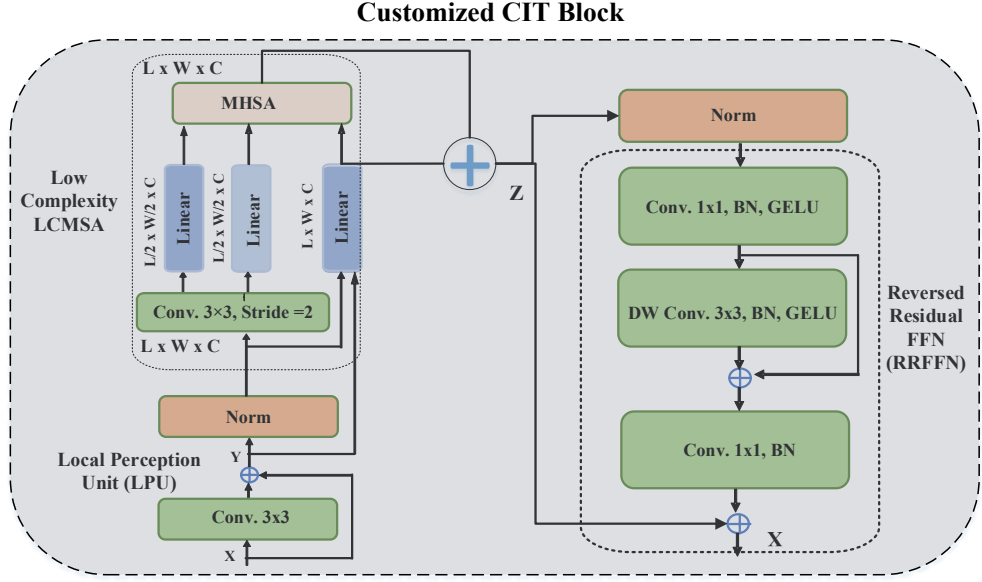

**Figure 3.** Structure of developed CIT Block introducing LPU, low-complexity MHSA, and RRFFN

The CIT block shows a significant capacity to capture both local and global features, as elaborated in Segment 3.2. Based on the framework of ViT, the current design will substantially improve the typical Multi-Head Self-Attention (MHSA) module by replacing it with a low-complexity version (LCMSA) and the typical MLP layer using a advanced Reversed Residual Feed Forward Network (RRFFN). The architecture consists of two sequential regularization sub-layers, followed by LCMSA and RRFFN, in a similar way. These structural improvements increase the model's capability to learn complex descriptions and enrich feature representations, thereby improving performance on challenging vision tasks. Additionally, the Local Perception Unit (LPU) is implemented within the CIT block for the enhanced system's representational capability. Finally, the developed CIT block is composed of three key components, as outlined below and in particular described in Eqs. 4–6.

$$y_k = LPU(x_k - 1) \qquad (4)$$

$$z_k = LCMHSA(LN(y_k)) + y_k \qquad (5)$$

$$xk_k = RRFFN(LN(k)) + z_k \qquad (6)$$

The output features of $i_{th}$ block from LPU and LCMHSA components are denoted as $y_i$ and $z_i$, respectively. NL is applied, and several CIT-blocks are arranged in order within each level to facilitate effective feature improvement and combination.

### i. Local Feature Extraction (LFE) Module

The Local Feature Extraction (LFE) module tackles invariance challenges in vision tasks, specifically by handling rotation and shift augmentations, and as a result, securing transformation invariance in the model's outputs [40]. Unlike earlier transformers, which employ unlimited positional encoding, assigning a positional value to each patch and thereby reducing invariance, ViTs focus on capturing local relationships [41] [42] and structural information [43] within patches. To address these limitations, the LPU is introduced to enable the extraction of fine-grained local information and enhance the model's ability to preserve spatial consistency. The input is '**x**' using sizes *L, W, C.*, where L×W and d denote the size of the input and feature dimension, respectively (Eq. 7).

$$(\pmb{x}) = Conv\_(\pmb{x}) + \pmb{x} \qquad (7)$$

### ii. Low-Complexity Multi-Head Self-Attention

The Multi-Head Self-Attention MHSA overcomes the weaknesses of the Single-Head Self-Attention (SHSA), which tends to emphasize limited pixels and may skip important regions. Self-Attention (SA) is the core component of ViTs and is famous for its ability to explicitly capture the relationships between elements within a sequence. Over time, numerous ViT architectures have been developed by refining the SA module to boost performance. Some models use condensed global attention, whereas others adopt sparse attention to learn dependencies in complex spatial images.

MSA employs the parallel stacking of multiple SA blocks to overcome this limitation, and it enhances the overall efficiency of the SA layer. This approach enables the model to embed each entity within a global context and capture its associations more comprehensively [44]. In the primary SA block, the input feature map '**x**' is linearly projected into key (k), query (q), and value (v) matrices. These projections assign a unique representation subspace to each component, allowing the MSA mechanism to learn diverse and complex interactions between sequence components. The mathematical formulation of this process is presented in Eqs. 8–10, with an overview provided in Eq. 3.

$$A(\mathbf{q}, \mathbf{k}, \mathbf{v}) = \sigma\left(\frac{\mathbf{q}.\mathbf{k}^T}{\sqrt{d_k}}\right) \qquad (8)$$

$$\mathbf{k}' = Conv.L(\mathbf{k}) \qquad (9)$$

$$\mathbf{v}' = Conv.L(\mathbf{v}) \qquad (10)$$

A 3×3 depthwise Conv-L (stride 2) is applied to the k and v matrices, reducing spatial dimensions and enhancing local feature extraction for attention.. Early transformer layers are replaced with convolutional blocks to downsample feature maps, which are then tokenized for high-level representation. The attention employs a 4×4 patch window with relative position bias B. The Low-Complexity Attention (LCA) mechanism is defined as:

$$LA(\mathbf{q}, \mathbf{k}', \mathbf{v}') = \sigma\left(\frac{q.k'^T}{\sqrt{d_k}} + B\right) v' \qquad (11)$$

The attention model, denoted as '$LA$' with activation 'σ,' uses q, v, and transposed key $k'^T$ as query, value, and key matrices in Eq. (11). Moreover, $\sqrt{d_k}$ aids as a scaling factor, along with '$d_k$' demonstrating the size of the key matrix. MSA comprises several SA blocks, each with learnable weight matrices for key, query, and value subspaces. During execution, the outputs from these blocks are combined and projected into the output space using a trainable parameter $Wo$. It enables the integration of learned contextual information. The proposed CIT architecture offers flexibility for fine-tuning in diverse detection tasks. The LCMHSA module employs multiple attention heads, each applying an LA function and generating outputs of size n × d/h. These are concatenated into a unified feature sequence of dimension n × d, where n denotes patch resolution and d the feature dimension, and $h$ represents the number of heads. The concatenation operation is denoted by 'cat'. This process is mathematically described in Eqs. (12–13) as follows:

$$LCMHSA(\mathbf{q}, \mathbf{k}', \mathbf{v}') = cat(h_1, h_2, \ldots \ldots h_h).W_0. \qquad (12)$$

$$h_i = LA(\mathbf{q_i}, \mathbf{k'_i}, \mathbf{v'_i}), \text{where } i = 1,2, \ldots \ldots, h \qquad (13).$$

### iii. Reversed Residual Feed-Forward Network

The anticipated Reversed Residual Feed-Forward Network (RRFFN) acts like a conventional residual block but enhances performance by repositioning the skip connection and integrating an expansion layer along with convolution operations. Usually, an FNN consists of two linear layers divided by a GELU activation function, which facilitates the extraction and integration of complex features, as expressed in Eq.14 [45]. In each encoder block, the FFN follows the Self-Attention (SA) block. To enhance efficiency, we integrate a combination of 3×3 and point-wise (1×1) convolutional layers within the transformer's FFN, effectively reducing the number of parameters and computational cost, while maintaining performance. Additionally, the proposed RRFFN introduces an inverted residual block with a skip connection, replacing the standard GELU activation and normalization layers. The inclusion of this shortcut connection aligns with the significances of classical residual networks, facilitating improved gradient flow across layers, as detailed in Eqs.15-16.

$$\text{FFN}(\mathbf{x}) = (b_2 + w_2 * \sigma_g(b_1 + w_1 * \mathrm{x})) \tag{14}$$

$$\text{RRFFN}(\mathrm{x}) = Conv(\mathcal{F}(\text{Conv}(\mathrm{x}))) \tag{15}$$

$$\mathcal{F}(\mathbf{x}) = Conv.L(\mathbf{x}) + \mathbf{x} \tag{16}$$

In addition, exploiting Conv-L enables an effective local information extraction. The GELU activation, denoted as $\sigma_g$, is applied with weights $w_1$ and $w_2$, whereas $b_1$ and $b_2$ correspond to biases Eq.14.

### 3.2.3 Residual Learning Blocks

We employed an innovative stacking strategy that integrates TL-based residual learning CNNs through 4 sequential K and L blocks for efficient feature learning Figure 4(A). In the L block, a 1×1 point-wise convolution enables feature interaction and transforms Conv-L outputs into distinct components. Both K and L blocks use a 3×3 kernel in Conv-L to establish local receptive fields. Combining these blocks in addition to the CIT architecture in the last stage enables the exploitation of different feature spaces. This configuration promotes the extraction of a broad spectrum of vital features. To strengthen the learning process, the number of channels is gradually increased from 64 to 256, ensuring a systematic and precise learning experience that enhances model performance.

The residual blocks leverage TL to generate additional feature maps, effectively expanding the channel diversity. These channels, originating from TL-based deep CNNs, strongly capture fine-grained variances and texture within brain tumor MRI scans. The residual architecture plays a vital role in identifying intricate features critical for differentiating contrast and texture patterns in BT MRIs. Additionally, the model's robustness is reinforced through the incorporation of fully connected and dropout layers, which preserve essential features and reduce the risk of overfitting.

$$y = T(x, \{w_i\}) + x \qquad (17)$$
$$y = T(x, \{w_i\}) + w_s x \qquad (18)$$

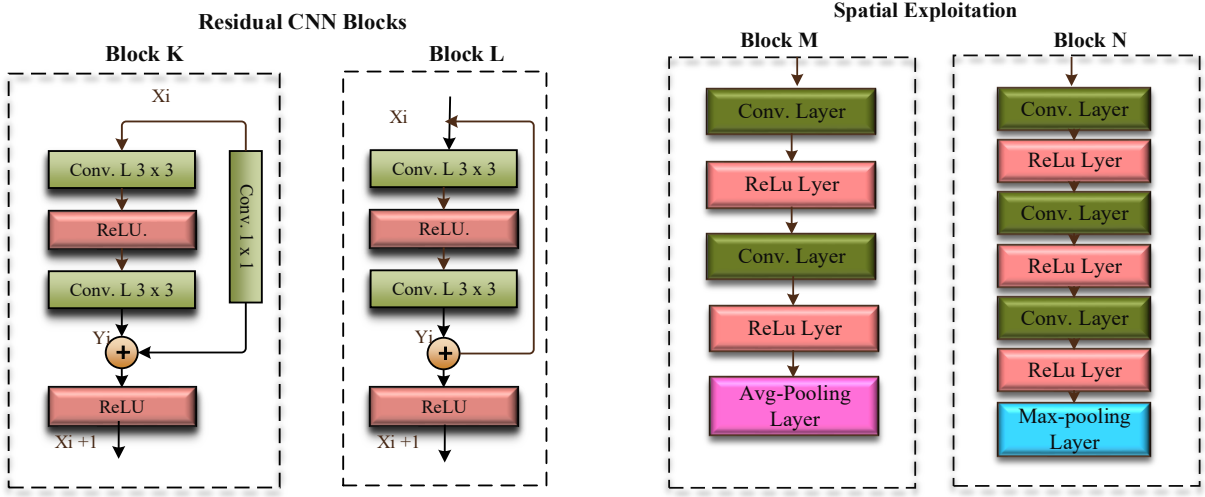

**Figure 4.** (A) Residual CNN Block Structure. (B) Spatial CNN Block Structure.

The residual block establishes a shortcut association between the input vector 'x' and the output vector 'y' of the target layers, as described in Eqs.. 17–18. Function $T(x, \{wi\})$ denotes the desired residual mapping. In a two-layer model (as exhibited in Figure 4(A)), $T = w_1 \sigma(w_1 x)$, where σ is the ReLU activation function. The $T + x$ operation is assisted by a shortcut connection with element-wise addition, followed by $\sigma(y)$. In cases where the dimensions of 'x' and 'T' do not match, a linear projection denoted by $w_s$ is applied through the shortcut path to align the dimensions and ensure compatibility.

### 3.2.4 Spatial Exploitation Blocks

Spatial blocks are employed to systematically capture inter-class contrast and local homogeneity variations in BTs. These blocks enable hierarchical feature extraction by utilizing multiple convolutional layers designed to capture spatial consistency and texture patterns. Each spatial block incorporates a squeezed 3×3 convolutional layer, subsequently batch normalization, an activation function, and multiple successive layers for efficient local feature extraction, as shown in Eqs.19–20 and Figure 4(B). This architectural design significantly enhances the network's ability to recognize diverse features within MIs, thereby improving detection accuracy and performance. Each spatial block includes 3×3 convolutional layers using a padding of 1, after that, a 2×2 max or Avg-pooling layer using a stride of 2. These layers process spatial data across multiple depths, using 3×3 filters in conjunction with both max and average pooling operations to ensure consistent feature extraction and accurate localization of fine-grained spatial boundaries [8][46].

The convolutional kernels preserve spatially homogeneous resolution, while the pooling layers capture high-contrast intensity features. Additionally, the pooling operation progressively reduces spatial resolution at each block, thereby enhancing robustness [47]. The architecture consists of five stacked spatial blocks, each comprising repeated sequences of convolutional and pooling layers, which together form a hierarchical feature extraction pipeline. This design enables the model to learn increasingly complex features at deeper layers while preserving essential information at the lower levels. The spatial blocks strengthen the network to learn effectively local, homogeneous, and correlated features, while the use of small kernels helps in minimizing the number of parameters and computational cost. Furthermore, the structure promotes better generalization through inherent regularization mechanisms.

### 3.2.5 Channel Enhancement (CE) Scheme

The proposed CE-RS-SBCIT framework effectively captures diverse image representations—including global context, texture details, and local features, by concatenating customized SBCIT, residual, and spatial feature maps Eq.19 [48]. The customized SBCIT utilizes a window-based self-attention mechanism, enabling it to detect subtle intensity contrast variations, while the novel R block focuses on extracting texture features that are potentially critical for identifying BTs. Additionally, the SBCIT channels are enriched with spatial and residual auxiliary feature maps generated through TL, thereby expanding the model's learning capacity. A specifically designed boosting strategy marked by a gradual increase in the number of channels facilitates comprehensive and fine-grained feature learning. Within this boosting framework, individual learners make decisions by analyzing diverse, image-specific patterns, ultimately enhancing the overall prediction quality.

$$\boldsymbol{X}_{Boosted} = (\boldsymbol{X}_R \,||\, \boldsymbol{X}_S \,||\, \boldsymbol{X}_{\text{SBCIT}}) \qquad (19)$$

In Eq.19, the CE-RS-SBCIT, denoted as R, S, and SBCIT, utilizes feature maps represented by $\boldsymbol{X}_R, \boldsymbol{X}_S$, and $\boldsymbol{X}_{\text{SBCIT}}$ respectively. Additionally, in some cases, SBCIT channels are combined with other generated residual and spatial learning channels using TL. The boosting process, denoted as (.), plays a vital role in learning diverse feature maps. Furthermore, the output is refined through a global average pooling layer, which reduces dimensionality while preserving spatial information. Finally, classification is performed using a fully connected layer alongside softmax activation, adapted to the dataset classes, including meningioma, glioma, pituitary tumors, and normal brain images.

### 3.2.6 Spatial Attention-based Approaches

We have implemented a pixel-attention-supported module to obtain class-related features at the spatial level. In this context, the spatially weighted attention block is recognized for its computational demands and its focus on the patterns of BTs Figure 5. The enhanced channels of different learners are combined and weighted due to the attention mechanism, as outlined in Eqs. 20-21.

$$Z_{SA-Out} = y_{pixels} \cdot z_{Enhanced} \quad (20)$$

$$z_{relu} = \sigma_1(y_X z_{Enhanced} + y_{SA} SA_{k,l} + b_{SA}) \quad (21)$$

This allows the network to focus on informative pixels while suppressing redundant ones. The figure illustrates element-wise addition and multiplication, along with the evaluation of various boosted channels and SA. The enhanced feature map input is represented as $z_{Enhanced}$. Refining the boosted feature map involves per-element addition among different pixel-weighted activations and the input, which results in the computation of $y_{pixels}$.

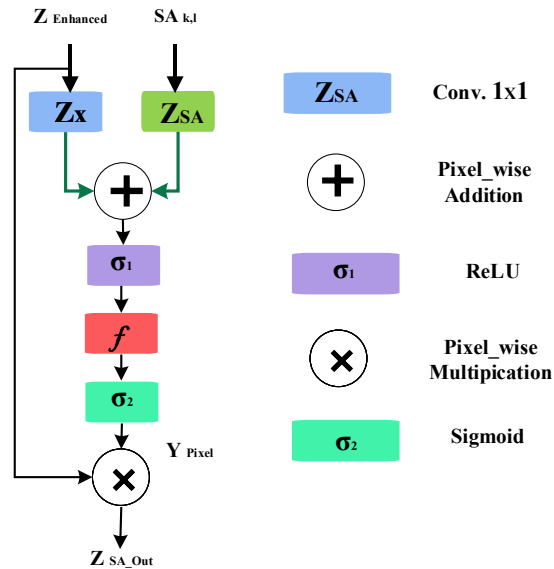

**Figure 5.** Spatial Attention Block pattern.

The enhanced channels are earlier element-wise multiplied with the spatial attention to generate the final refined feature map, $Z_{SA-Out}$. Lastly, the sequential weighted output from the SA block is passed through fully connected layers, as illustrated in Eqs. 20-22. This procedure boosts the label-specific feature map, amplifies the differentiating pixel impact, and boosts overall performance of the model.

$$y_{pixels} = \sigma_2(f(z_{relu}) + b_f) \qquad (22)$$

$$z = \sum_a^A \sum_b^B V_a Z_{SA-Out} \qquad (23)$$

$$\sigma(x) = \frac{e^{x_i}}{\sum_{i=1}^d e^{x_d}} \qquad (24)$$

In Eq.20, $z_{Enhanced}$ represents the input map, and $y_{pixels}$ denotes the weighted-pixel in the limits of [0, 1]. The final output, $Z_{SA-Out}$, emphasizes affected regions while suppressing irrelevant features. Eqs. (21–22) define activations $\sigma_1$, $\sigma_2$, biases $b_{SA}$, $b_f$, and transformations $y_X$, $y_{SA}$, and f. Eq. (22–23) represent the neuron count $V_a$ and the softmax activation σ

### 3.3 Employment of Existing CNNs and ViTs

In this study, we employed advanced ViT and CNN models for performance comparison. To effectively classify BTs in MRI images, we employ multiple deep CNN and ViT architectures across diverse datasets [49]. Different CNN and ViT models, incorporating ResNet-18, VGG-16, EfficientNetB0, InceptionResNetV2, GoogleNet, InceptionV3, DarkNet-53, SqueezeNet, MobileNet, DarkNet-19, ViT, Tiny ViT, MobileViT, Dual-ViT, Cross ViT, SwinT, LeViT (Hybrid), TinyViT++, SwinV2, and LocalViT, are used for BT analysis. CNNs have demonstrated proficiency in detecting BTs in the medical field [50]. Deep CNNs, with varying depths and architectures, are explicitly developed for exploring different types of BTs. ViT models have shown enhanced accuracy, even on smaller datasets. To enhance local feature extraction in ViTs, researchers employ depthwise convolutions and positional encoding. LeViT incorporates convolutional layers to learn spatial and low-level representations efficiently.

# 4. Implementation details
## 4.1 Dataset

Accurate classification of brain tumors from MRI is a key challenge in MI analysis, primarily due to the lack of high-quality, balanced, and diverse datasets BRISC [49], Kaggle [51], Br35H [52], and figshare [53]. In this work, we have collected the strengthened dataset, which consists of 16,287 contrast-enhanced T1-weighted MRI scans, annotated by expert radiologists and specialists. The dataset contains three major tumor types: glioma, meningioma, and pituitary tumors, along with non-tumorous cases. All samples are accompanied by high-resolution labels and categorized across axial, sagittal, and coronal imaging planes, enabling robust model development and cross-view generalization. The dataset contains MRI images of healthy humans and three distinct kinds of BTs. MRI images of four classes are collected from open-source depositories [49], [51], [52], [53].

We gathered 3,464 MRI normal images, 3,948 glioma tumor images, 4,217 meningioma tumor images, and 4,658 pituitary tumor images from standard repositories. Consequently, the obtained dataset is imbalanced, as illustrated in Table 1. The dataset consists of MRI images of both normal brain tissue and various categories of tumors. Input images were resized to meet the input size requirements of the DCNNs. Examples of images from the four classes are provided in Figure 6.

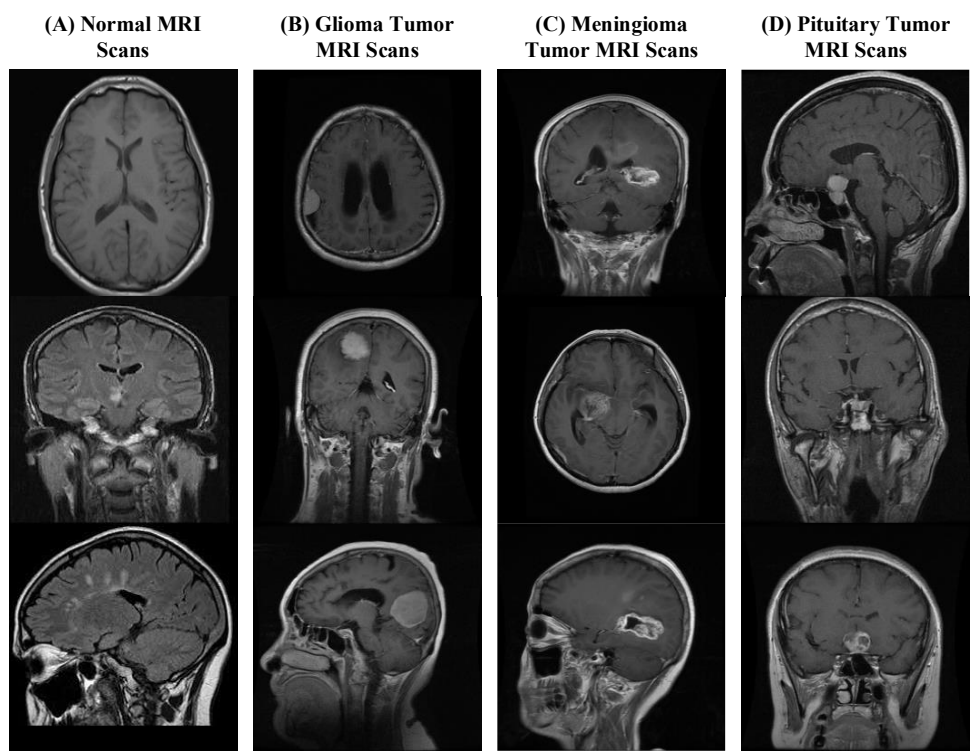

**Figure 6.** Sample MRI scans of normal and different classes of tumors

**Table 2.** Details of the collected data set of MRI scans of normal and diverse types of tumors.

|  | **Glioma** | **Meningioma** | **Pituitary** | **Normal** | **Total** |
|---|---|---|---|---|---|
| **Training (70%)** | 2763 | 2951 | 3260 | 1040 | 11398 |
| **Testing (30%)** | 1185 | 1266 | 1398 | 2424 | 4889 |
| **Total (100%)** | 3948 | 4217 | 4658 | 3464 | 16287 |

### 4.2 Experimental Setup and Implementation

The CE-RS-SBCIT model and recent ViTs/CNNs both used Adam as an optimizer for training. The learning rate was set to $10^{-3}$ with an 85% decay after every 20 epochs. The class imbalance issue has been tackled by using the cross-entropy loss function for classification, with a batch size of 16 and a 0.3 dropout rate at the output layer. All experimentations were carried out on a system running MATLAB 2024-b, featuring an Intel Core i7-10G processor, 32 GB of RAM, and an NVIDIA GeForce RTX-3060 GPU with 32GB of RAM.

A total of 16,287 subjects from open-source data were used in hold-out cross-validation, with 20% reserved for validation in each iteration. Performance was evaluated using accuracy, sensitivity/recall, precision, F1-score, ROC/PR curves, and AUC (Eqs. 22–26). As shown in Eq. (26), the standard error (S.E.) of sensitivity within a 95% confidence interval (CI) was computed for statistical analysis using z = 1.96 to maximize true positives (TPs) and minimize false negatives (FNs) in BT detection.

$$\text{Acc.} = ((TP + TN)/(TP + TN + FN + FP)) \times 100 \quad (22)$$

$$\text{Sen.} = TP/(TP + FN) \quad (23)$$

$$\text{Pre.} = TN/(TN + FP) \quad (24)$$

$$F1 - \text{Score} = (2 \times (\text{Pre.} \times \text{Sen.}))/(\text{Pre.} + \text{Sen.}) \quad (25)$$

$$CI = z\sqrt{((error(1 - error))/(total\ BT\ Samples))} \quad (26)$$

### 5. Results and Discussion

This section presents the experimental evaluation followed by ablation studies conducted to validate the effectiveness of the integrated SBCIT and CE-RS-SBCIT techniques. The proposed CE-RS-SBCIT technique is benchmarked against existing ViT, CNN, and hybrid ViT-CNN architectures using the Kaggle brain tumor dataset. Performance has been assessed using accuracy, sensitivity, precision, F1-score, and AUC (Table 3). The multi-class classification task involved categorizing MRI scans into meningioma, glioma, pituitary, and normal classes. The SBCIT backbone outperformed competing models, attaining 96.8% accuracy, 96.7% sensitivity, 96.4% precision, and a 96.55% F1-score, thereby demonstrating its superior discriminative capability in brain tumor diagnosis.

Integrating CNNs with transformers in the architecture significantly enhanced performance on the challenging dataset. In addition, the CE-RS-SBCIT technique outperformed existing ViTs/CNNs, achieving an accuracy of 98.3%, sensitivity of 98.43%, precision of 98.08%, and an F1-score of 98.25%. Additionally, it attained remarkable PR-AUC (0.9875) and ROC-AUC (0.9894) scores. The outcomes confirmed the efficacy of the CE approach within the three-stream structure, which incorporates SBCIT, spatial exploitation, residual learning, and CNN, in improving feature encoding and generalization. The CE-RS-SBCIT model beats recent models on the open source dataset and is also compared with the customized residual, spatial CNNs, and the proposed SBCIT, as illustrated in the confusion matrix presented in Figure 7.

The confusion matrix of the proposed CE-RS-SBCIT technique indicates that 1,158 glioma cases have been correctly identified, with 2% of MRI images misclassified across the normal, meningioma, and pituitary categories. Specifically, 4 pituitary cases are incorrectly predicted as glioma, 8 as meningioma, and 7 as normal, while the remaining pituitary cases are classified accurately. For meningioma, 97.9% of cases are recognized correctly, with only 2.1% misclassified. Similarly, 97.9% of normal cases were classified accurately, whereas 2.1% is incorrectly assigned to glioma, meningioma, or pituitary classes.

In general, the total number of cases can be correctly classified, and only 1.7% of MRI images can be classified incorrectly. Table 3 presents the performance summary of the proposed approach in multi-class analysis. Finally, the CE-RS-SBCIT approach outperforms SBCIT with the same number of sources on true positives (TP), and a decrease in false negatives (FN) as well as false positives (FP) when compared to existing ViT/CNN-based approaches and the SBCIT. Assessing the computational performance of algorithms is imperative, especially when dealing with large datasets. We assess the computational complexity of our light-weight, skip-connection custom-made SBCIT model for precise BT classification using MRI data, and the training complexity is illustrated in Figure 7.

Particularly, the lightweight SBCIT model requires fewer computing resources and less training time compared to other ViT models, while attaining optimum performance with an efficient architecture. Additionally, this enables quicker and more efficient convergence to the optimal result, as well as more cost-effective training, making it applicable for resource-constrained hardware and thereby improving BT categorization accuracy. In contrast, the performance of the recent hybrid (CNN-ViT) model varies at the time of convergence, striving to find the best solution.

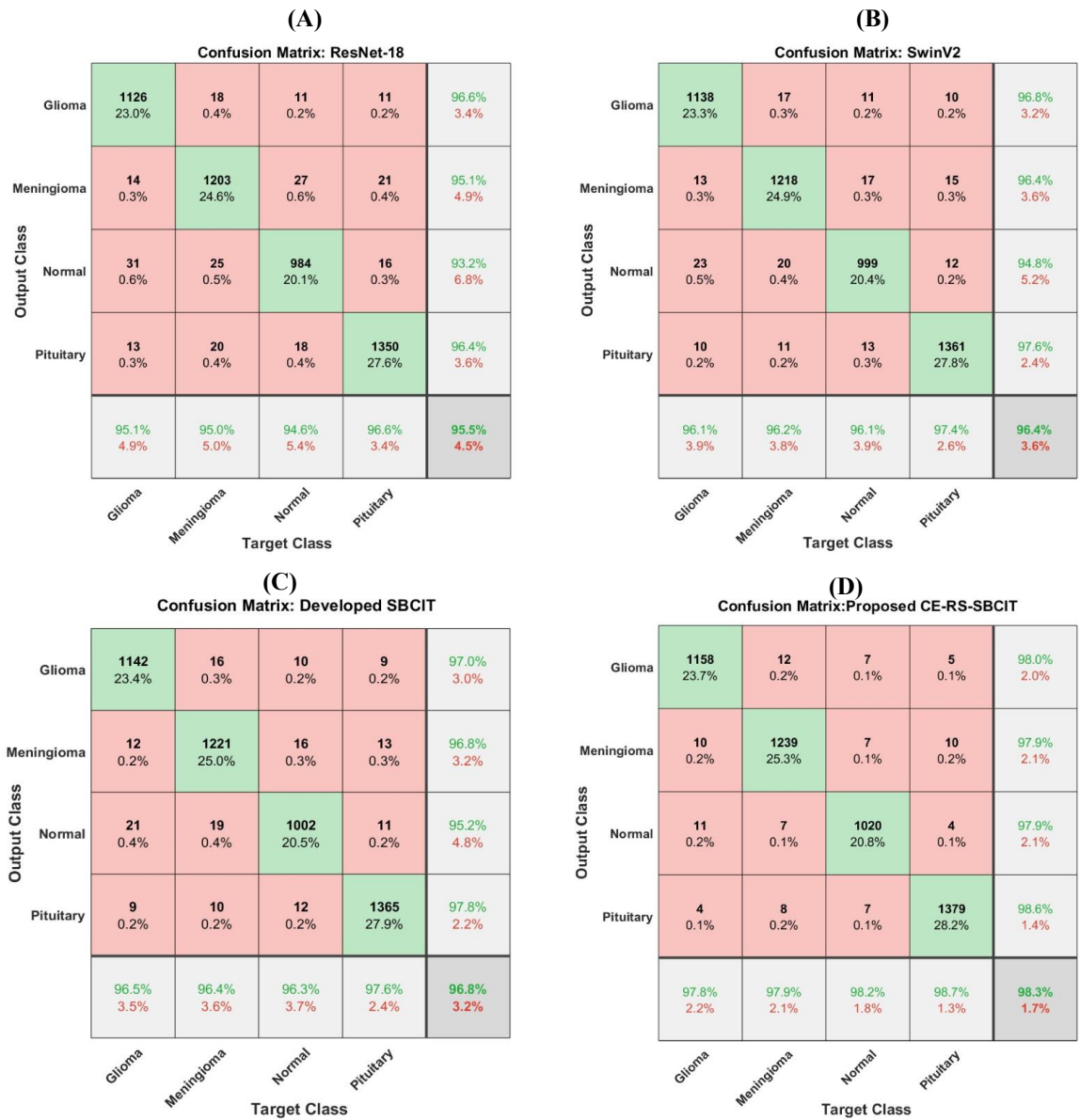

**Figure 7.** Confusion Matrix of the proposed techniques and existing ViTs/CNNs.

## 5.1 Comparison with the Existing ViTs/CNNs

This part contains a significant comparison of our given CE-RS-SBCIT and SBCIT techniques to detect BT with other models, performing best on open-source datasets in Kaggle. Different CNN and ViT models are applied in BT analysis, such as EfficientNetB0, InceptionResNetV2, GoogleNet, InceptionV3, DarkNet-53, SqueezeNet, MobileNet V4, DarkNet-19, VGG-16/19, ShuffleNet, ResNet-50, Xception, ResNet-18, LocalViT, Cross ViT, Dual ViT, Tiny ViT, and Hybrid ViT. Table 3 demonstrates the performance of our model in the area of multi-class classification, which supports the fact that it is a powerful model when compared to other great methods of DL applied to a similar dataset.

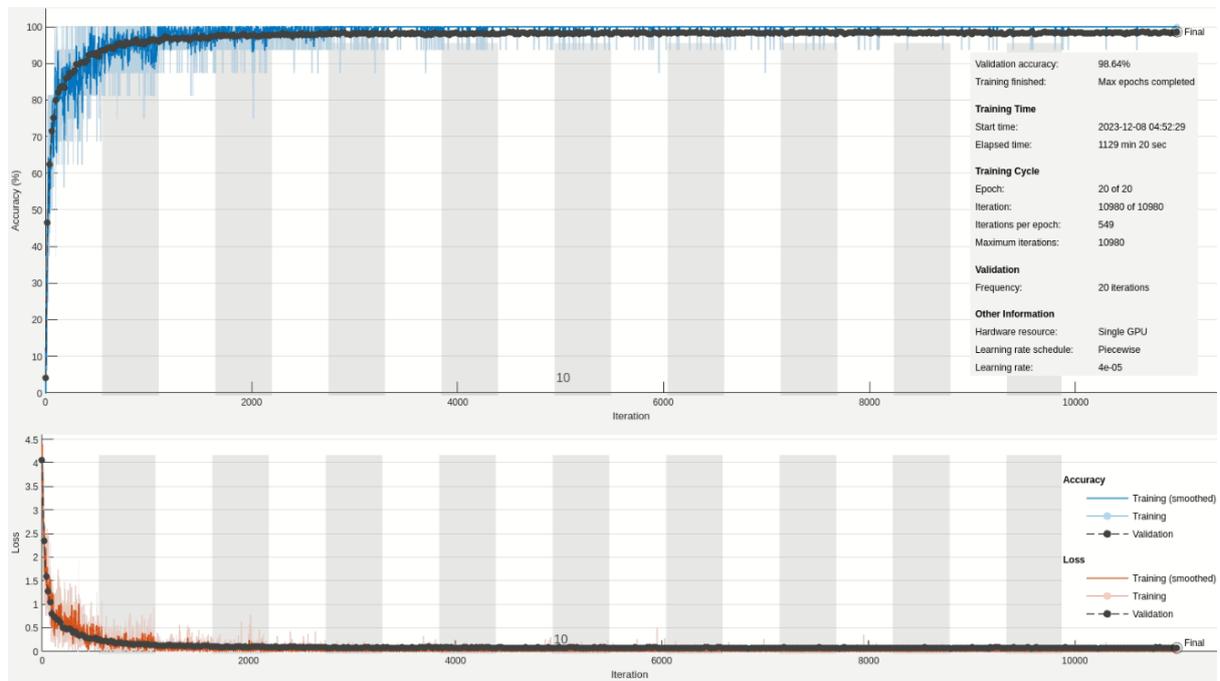

**Figure 1.** Training plots; proposed CE-RS-SBCIT.

The developed technique exhibits significant progress over recent CNNs, with performance gains ranging from 3.3% to 6.9% in accuracy (Acc.), 3.33% to 7.03% in sensitivity (Sen.), 3.28% to 6.88% in precision (Pre), and 3.3% to 6.95% in the F1-score (Figure 9(C)). Furthermore, the proposed method outperforms existing global receptive learning ViTs, achieving improvements of 1.9% to 4.0% in accuracy (Acc.), 1.93% to 4.03% in sensitivity (Sen.), 1.98% to 3.98% in precision (Pre.), and 1.95% to 4.0% in the F1-score (Figure 9(D)). A comparative study of recent techniques used on Kaggle BTs is presented in Table 3. Our findings demonstrate the notable performance of the proposed CE-RS-SBCIT, achieving the maximum results in multiclass classification tasks. Unlike their complex model, which involves separate classifiers, our lightweight approach facilitates real-world deployment.

Table 3. Comparative exploration of the proposed methodology with current CNNs/ViTs.

| Model | Acc. | Sen. | Pre. | F1-Score ± S.E. | FLOPs (G) | Inference (ms) | Time/Epoch (min) | Train Time (min) |
|---|---|---|---|---|---|---|---|---|
| **Existing CNNs** | | | | | | | | |
| EfficientNetB0 | 91.4 | 91.2 | 91.4 | 91.30 ± 0.045 | ~0.39 | ~5.4 | 0.34 | 10.2 |
| InceptionResNetV2 | 91.9 | 91.6 | 91.9 | 91.75 ± 0.044 | ~15.2 | ~18.7 | 0.88 | 26.4 |
| GoogleNet | 92.3 | 92.3 | 92.3 | 92.30 ± 0.042 | ~1.5 | ~6.1 | 0.42 | 12.6 |
| InceptionV3 | 92.6 | 92.4 | 92.7 | 92.55 ± 0.043 | ~5.7 | ~10.3 | 0.59 | 17.7 |
| DarkNet-53 | 93 | 92.8 | 93.3 | 93.05 ± 0.041 | ~18.7 | ~21.5 | 1.02 | 30.6 |
| SqueezeNet | 93.5 | 93.3 | 93.5 | 93.40 ± 0.039 | ~0.83 | ~4.2 | 0.3 | 9 |
| MobileNet V4 | 93.8 | 93.5 | 93.8 | 93.65 ± 0.038 | ~0.60 | ~3.9 | 0.31 | 9.3 |
| DarkNet-19 | 94.3 | 94.1 | 94.4 | 94.25 ± 0.038 | ~7.5 | ~20.3 | 0.83 | 24.9 |
| VGG-16 | 94.7 | 94.5 | 94.8 | 94.65 ± 0.037 | ~7.2 | ~12.1 | 0.77 | 23.1 |
| ResNet-18 | 95 | 94.8 | 95.1 | 94.95 ± 0.036 | ~3.8 | ~7.9 | 0.48 | 14.4 |
| **Existing ViTs** | | | | | | | | |
| CrossViT | 94.3 | 94.1 | 94.4 | 94.25 ± 0.038 | ~5.6 | ~15.9 | 0.58 | 17.4 |
| LocalViT | 94.1 | 93.9 | 94.4 | 94.15 ± 0.040 | ~6.8 | ~17.2 | 0.6 | 18 |
| TinyViT | 94.5 | 94.2 | 94.8 | 94.45 ± 0.041 | ~7.3 | ~16.9 | 0.62 | 18.6 |
| MobileViT | 94.8 | 94.6 | 94.9 | 94.75 ± 0.039 | ~1.1 | ~6.3 | 0.5 | 15 |
| SwinT | 95.7 | 95.5 | 95.8 | 95.65 ± 0.034 | ~8.5 | ~19.1 | 0.85 | 25.5 |
| **Hybrid ViTs** | | | | | | | | |
| Dual-ViT | 95.4 | 95.3 | 95.7 | 95.50 ± 0.035 | ~1.9 | ~7.1 | 0.54 | 16.2 |
| LeViT (Hybrid) | 96 | 95.8 | 96 | 95.90 ± 0.033 | ~4.5 | ~10.9 | 0.61 | 18.3 |
| ViTAR | 96.2 | 95.9 | 96.2 | 96.05 ± 0.033 | ~4.1 | ~9.7 | 0.59 | 17.7 |
| SwinV2 | 96.4 | 96.1 | 96.5 | 96.30 ± 0.032 | ~1.6 | ~6.7 | 0.52 | 15.6 |
| **Proposed Setup** | | | | | | | | |
| Developed (CIT) | 96.5 | 96.2 | 96.6 | 96.40 ± 0.031 | ~4.9 | ~11.2 | 0.63 | 18.9 |
| Developed (SBCIT) | 96.8 | 96.4 | 96.7 | 96.55 ± 0.031 | ~2.3 | ~7.5 | 0.48 | 14.4 |
| Proposed SBCIT-SA | 97.1 | 96.8 | 97.1 | 96.95 ± 0.029 | ~2.3 | ~7.5 | 0.48 | 14.4 |
| Proposed(CE-S+SBCIT | 97.9 | 97.4 | 97.9 | 97.65 ± 0.027 | ~2.3 | ~7.5 | 0.48 | 14.4 |
| Proposed CE-R-SBCIT | 98.1 | 97.8 | 98.4 | 98.10 ± 0.025 | ~2.3 | ~7.5 | 0.48 | 14.4 |
| Proposed CE-RS-SBCIT | 98.3 | 98.08 | 98.43 | 98.25 ± 0.023 | ~2.3 | ~7.5 | 0.48 | 14.4 |

**Table 4.** Comparative analysis with the recent studies.

| Author/Model | Architecture | Data Set Size (MRI) | Tumor Classes | Accuracy. | Limitation |
|---|---|---|---|---|---|
| H. Khan et al. (2020) [54] | VGG-16, ResNet-50, Inception-v3 | 253 images | 2 (Glioma, Meningioma) | VGG: 96.10% ResNet: 89.10% Inception-V3: 75.30% | • Comparatively Small dataset<br>• Binary classification<br>• Limited generalizability |
| S. Pokhrel et al. (2022) [55] | MobileNetV2, V3, VGG (16,19) VGG19 | 3000 images | 2 (Tumor, No-tumor) | MobileNet (V2,V3) (94.17%,94.83%) VGG (16,19): (97.2%, 97.5%) | • Small dataset<br>• No data augmentation<br>• Binary classification |
| Pillai et al. (2023) [56] | VGG16, ResNet50, and Inception V3 TL models with fine-tuning layers | 251 images | 2 (Tumor, No-tumor) | VGG16: 91.58% ResNet50: 81.94% InceptionV3: 63.86% | • Focused on binary classification<br>• Relatively small dataset |
| N. Shamshad et al. (2024) [57] | TL with VGG-16, ResNet-50, and MobileNet on MRI images using CNNs | 256 images | 2 (Tumor, No-tumor) | VGG-16: 97.2% ResNet-50: 96.0% MobileNet: 87% | • Comparatively Small dataset<br>• Binary classification |
| B. Sandhiya et al. (2024) [58] | Inception V3 and DenseNet201 as backbone | Dataset 1: 3264 images Dataset 2: 3064 images | 4(Glioma, Meningioma, Pituitary, Normal)<br><br>3(Glioma, Meningioma, Pituitary) | Dataset 1: 97.97% and Dataset 2: 98.21% | Data Set 1:<br>• Relatively Small dataset<br><br>Data Set 2:<br>• Small dataset<br>• 3 classes only, and healthy instances are not included |
| H. Mzoughi, I et al. (2025) [37] | Vision Transformer (ViT) | 3064 MRI images | 3(Glioma, Meningioma, Pituitary) | Vision Transformer (ViT): 91.61% | • Small dataset<br>• 3 classes only, and healthy instances are not included |
| M. Aloraini et al.(2023) [38] | Hybrid Transformer-Enhanced CNN | BraTS 2018 dataset | High-Grade and Low-Grade Gliomas | TECNN: 96.75% | • Focused on binary classification<br>• Relatively small dataset |
| S. Hong, J. et al. (2024) [39] | Modified VIT-B/16 | 7023 MRI images | 3(Glioma, Meningioma, Pituitary) | Modified VIT-B/16: 91.36% | • Small dataset<br>• 3 classes only, and healthy instances are not included |
| **Proposed CE-RS-SBCIT** | **CE-RS-SBCIT** | **16287 images** | **4 (Glioma, Meningioma, Pituitary, Normal)** | **98.30%** | - |

## 5.2 Ablation Study for Assessing the Proficiency of the Proposed Technique

In CE-RS-SBCIT, we fine-tuned the transformer and convolutional blocks using the new SBCIT, TL-supported residual learning, and spatial exploitation-based CNNs. We performed an ablation study using various block configurations to evaluate their effect on performance and effectiveness, with detailed results presented in Table 3. Traditional CNNs are limited in their ability to capture global receptive fields, whereas ViT models struggle with local feature abstraction. Furthermore, the lack of preliminary CNN methods in the backbone led to a slight decrease in the model's ability to generalize on test data, thereby improving performance, though at the expense of increased computational cost. To overcome this, we developed a hybrid approach by replacing the traditional blocks with STEM-CNN blocks and custom-made lightweight transformer blocks. The proposed lightweight CIT demonstrates significant improvements over the hybrid CNN-ViTs (LeViT), with accuracy (Acc.) increasing from 96% to 96.5%, sensitivity (Sen.) rising from 95.80% to 96.20%, precision (Pre.) improving from 96.0% to 96.6%, and F1-score advancing from 95.90% to 96.40%.

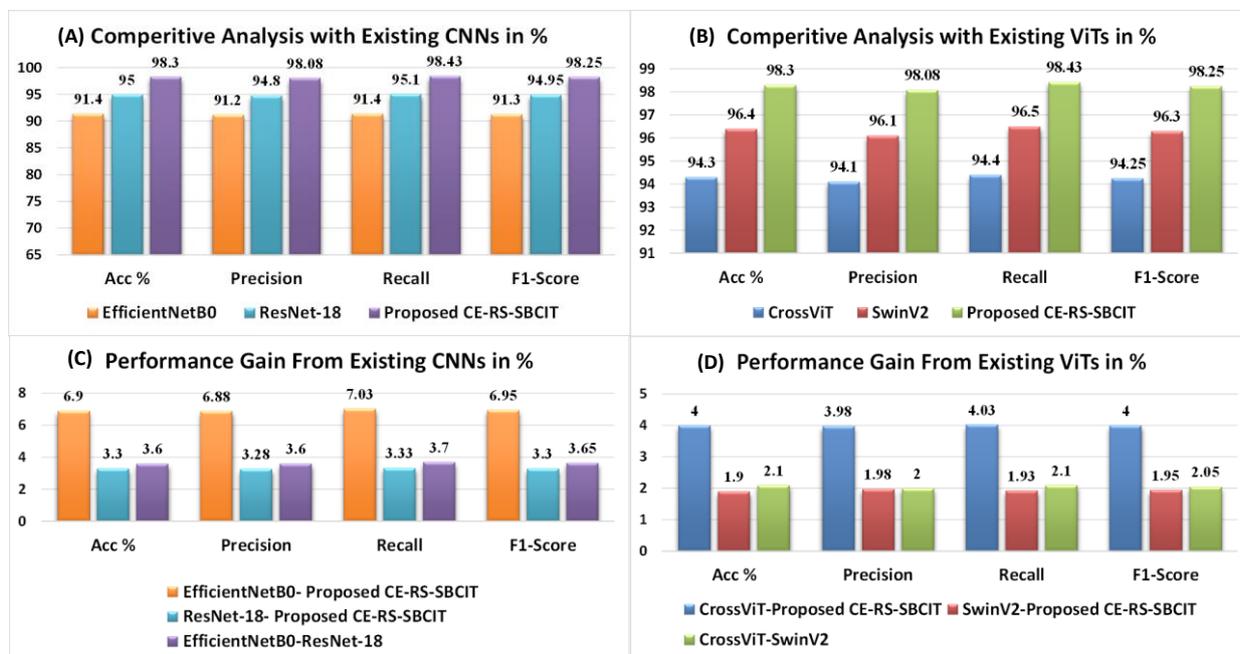

**Figure 9.** The proposed CE-RS-SBCIT performance analysis.

Furthermore, the integration of systematic smoothing and boundary operations within the customized CIT model, using the BT dataset, leads to improved performance with an accuracy of 96.8% and an F1-score of 96.55%. This approach demonstrates improvements in classification, with gains in accuracy (0.4-1.4%), sensitivity (0.3-1.1%), precision (0.3-1.0%), and F1-score (0.25-1.5%) compared to transformer-based models. The proposed CE-RS-SBCIT utilizes light-weight approaches, a two-stream, residual, and spatial learning to streamline the backbone network and mitigate overfitting. The overall performance improvement of the proposed CE-RS-SBCIT model is illustrated in Figure 8. Additionally, the combination of SBCIT boosting and residual learning and spatial exploitation CNN, integrated into the CE-RS-SBCIT, outperforms all state-of-the-art ViT and CNN methods, as demonstrated by the performance metrics, PCA-based analysis, and ROC-PR curves (Table 3, Table 4, and Figures 7-12).

### 5.3  PR/ROC curves

The study employs diagnostic accuracy, PR, and ROC curves to assess the performance of the CE-RS-SBCIT model. Detection rate curves are employed to quantitatively evaluate the model's ability to discriminate across different threshold setups, specifically examining its generalization among the normal class and others. The PR/ROC curves present an instant analysis of the precise prediction rate of the normal class compared to other BT classes.

Figure 10 illustrates key performance metrics, including PR-AUC, ROC-Accuracy, Sensitivity, Precision, and F1-Score. The model demonstrates significant performance gains over existing CNN methods, improving accuracy, sensitivity, precision, and F1-score across both minimum and maximum thresholds. Additionally, CE-RS-SBCIT outperforms existing ViT methods, showing superior AUC diagnostic accuracy across active learning techniques. The proposed approach achieves the highest area under the curve (AUC), signifying exceptional classification performance. PR and ROC curves are plotted by comparing predicted likelihoods to ground-truth labels, regarding performance results précised in Table 3. at 20% label availability. CE-RS-SBCIT outperforms all other existing learning methodswith a PR-AUC of 0.9875 and a ROC-AUC of 0.9894. As shown in Figure 10, CE-RS-SBCIT exhibits the best ROC curves and the maximum AUC compared to other CNN and ViT models. The blue curve demonstrates superior ROC performance, which is positioned nearby to the top-left corner, whereas better PR performance is represented by the same blue curve near the top-right corner. Consistently, in the ablation experiments depicted in Figure 10, CE-RS-SBCIT produces the most favorable ROC and PR curves among all tested versions, with the blue curve exhibiting superior performance in both metrics.

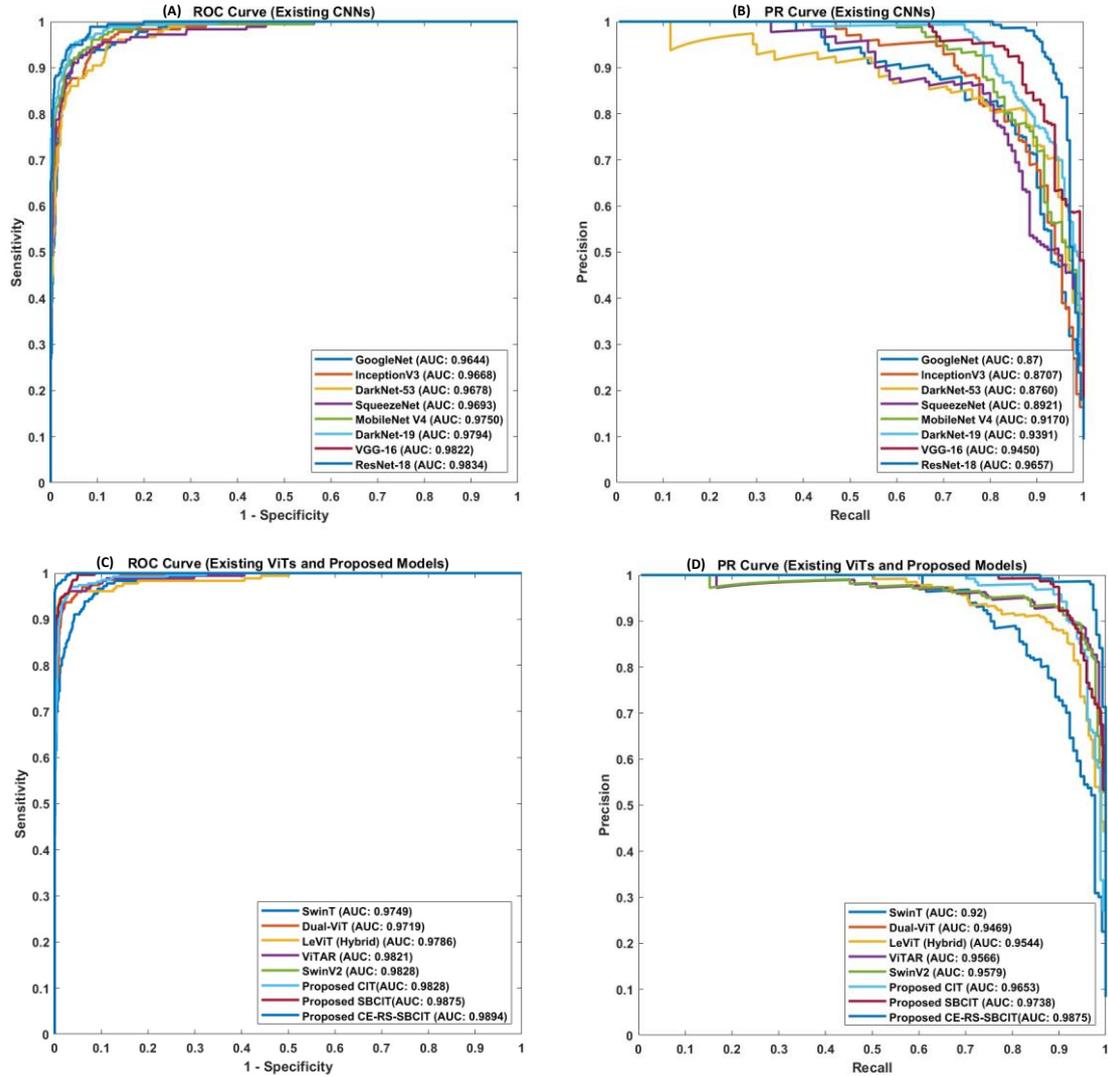

**Figure 10.** Discriminating rate exploration of the proposed model and current CNNs/ViTs

## 5.4 Feature Discrimination Analysis

The PCA projections for CE-RS-SBCIT, LeViT (Hybrid), and ResNet18, using Kaggle datasets, are shown in Figure 11. The plots clearly demonstrate that CE-RS-SBCIT achieves highly separated features in the final stages, allowing for easy distinction between the Glioma, Meningioma, Pituitary, and Normal classes. The effectiveness of the proposed hybrid CE-RS-SBCIT diverse learning is evident when comparing the hypergraph embeddings of the proposed model with those of LeViT (Hybrid) and ResNet18. In the PCA projections, the data points are color-coded according to their diagnosis: red for Glioma, magenta for Meningioma, green for Pituitary, and blue for Normal. CE-RS-SBCIT hypergraph embeddings (rightmost) exhibit clearer separability than other models (left), highlighting stronger feature representation and analysis performance.

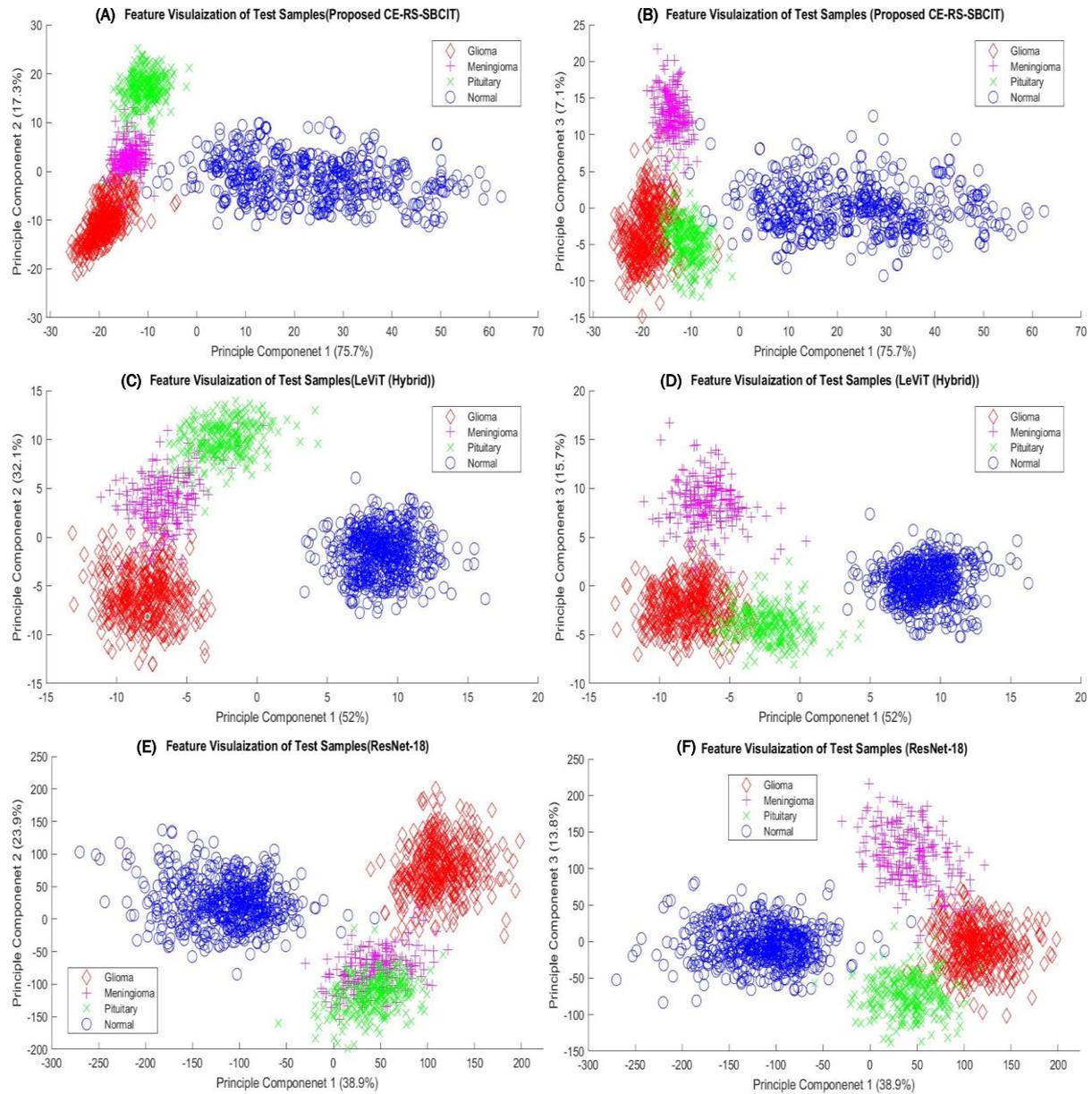

**Figure 11.** Feature Space Visualization of the proposed model on test samples

## 5.5 The Proposed Framework Significance

The significance of the proposed framework lies in its ability to unify multiple architectural innovations to achieve robust and generalizable brain tumor diagnosis from MRI scans:

- The proposed CE-RS-SBCIT network fuses residual learning, spatial learning, and CE by concatenating diverse maps to learn a multi-level representation. Integration of residual and

- spatial learning with channel enhancement enables the model to capture heterogeneous tumor characteristics at multiple representational levels, improving generalization across patient populations.
- Smoothing and boundary-driven CNN–Transformer fusion jointly models local textural variations and global contextual dependencies, ensuring resilience against morphological irregularities and subtle contrast variations. Systematic multi-level fusion of local and global cues bridges the gap between CNNs' localized feature strength and Transformers' global dependency modeling, producing a highly discriminative feature space.
- Region operations suppress noise and improve robustness, while edge operations sharpen tumor borders, strengthening the discrimination of boundary-adjacent lesions and enhancing tumor delineation in low-contrast MRI scans.
- Lightweight convolutional and transformer blocks reduce computational overhead and model size, making the framework suitable for real-time or resource-constrained clinical environments. Moreover, Local perception operations preserve spatial invariance, while multi-head self-attention refinements capture global contextual relations without excessive computational cost. Finally, Inverse residual connections improve gradient flow in deeper networks and strengthen fine-grained local texture encoding critical for discriminating tumor boundaries.
- TL–based residual and spatial feature blocks enrich the representation space, alleviating challenges of limited and imbalanced MRI datasets and enhancing classification reliability. Moreover, CE mechanisms amplify highly discriminative channels while suppressing redundant ones, leading to compact yet informative representations.
- Spatial attention refinement selectively emphasizes tumor-relevant pixels, improving inter-class separability and minimizing misclassification of ambiguous cases. The extensive evaluation across diverse datasets demonstrates strong generalization, highlighting the model's potential for deployment in clinical decision support systems.

# 6. Conclusion

This study introduces a hybrid diagnostic framework, CE-RS-SBCIT, that integrates convolutional and transformer paradigms for automated brain tumor classification from MRI scans. The design systematically addresses clinical challenges such as subtle contrast shifts, morphological heterogeneity, and textural ambiguity by combining local features encoding with global dependency modeling. At the initial stage, a STEM convolution extracts locally correlated features and forwards them for transformer tokenization, ensuring efficient local-to-global alignment and reduced computational complexity. The customized CNN-Transformer stream incorporates a smoothing operation to reduce noise and improve robustness, while boundary operations sharpen tumor borders, enhancing the discrimination of boundary-adjacent lesions. Residual CNNs, enhanced by inverse residual concepts, stabilize gradient propagation and strengthen deep feature extraction, whereas spatial CNNs emphasize fine-grained textural and contrast cues. Lightweight transformers and attention blocks capture global dependencies with reduced overhead, complemented by channel enhancement that amplifies discriminative responses while suppressing redundancy. Transfer learning and augmentation mitigate dataset imbalance and improve generalization, and a novel spatial attention mechanism selectively emphasizes tumor-relevant pixels to refine inter-class separability. Extensive evaluation against eight state-of-the-art frameworks confirms consistent improvements, surpassing the best hybrid baseline (LeNet-CNN–ViT) with accuracy +2.30%, precision +2.28%, sensitivity +2.43%, and F1-score +2.35%. Collectively, these innovations deliver a computationally efficient, robust, and generalizable solution for brain tumor diagnosis. Future research will extend this framework through multi-modal integration of imaging and clinical data, while generative modeling will address data scarcity. Incorporating explainable AI will further enhance interpretability and clinical adoption.